\documentclass[runningheads]{llncs}
\usepackage{booktabs}
\usepackage{amsmath}
\usepackage{graphicx}
\usepackage{hyperref}
\usepackage{url}
\usepackage{enumitem} 
\usepackage[dvipsnames]{xcolor}
\graphicspath{{images/}}  

\renewcommand\UrlFont{\color{blue}\rmfamily}

\begin{document}
\title{Top-1 CORSMAL Challenge 2020 submission: Filling mass estimation using multi-modal observations of human-robot handovers}
\titlerunning{Filling mass estimation using multi-modal observations}

\author{\mbox{Vladimir Iashin\inst{1}\orcidID{0000-0001-8879-587X}\thanks{All authors have contributed equally} \and
Francesca Palermo\inst{2}\orcidID{0000-0002-1862-4853}}$^{\star}$ \and \\
\!G\"okhan\,Solak\inst{2}\orcidID{0000-0001-6342-1345}$^{\star}$ \and
\!Claudio\,Coppola\inst{2}\orcidID{0000-0002-3835-9268}$^{\star}$}

\authorrunning{V. Iashin et al.}
\institute{
Computer Vision Group, Information Technology and Communication Sciences, Tampere University, Tampere, 33720, Finland
\\\email{vladimir.iashin@tuni.fi}\\
\and
ARQ (Advanced Robotics at Queen Mary), School of Electronic Engineering and Computer Science, Queen Mary University of London, London, E14NS, UK
\email{\{f.palermo, g.solak, c.coppola\}@qmul.ac.uk}}
\maketitle
\begin{abstract}
Human-robot object handover is a key skill for the future of human-robot collaboration. CORSMAL 2020 Challenge focuses on the perception part of this problem: the robot needs to estimate the filling mass of a container held by a human. Although there are powerful methods in image processing and audio processing individually, answering such a problem requires processing data from multiple sensors together. The appearance of the container, the sound of the filling, and the depth data provide essential information. We propose a multi-modal method to predict three key indicators of the filling mass: filling type, filling level, and container capacity. These indicators are then combined to estimate the filling mass of a container. Our method obtained Top-1 overall performance among all submissions to CORSMAL 2020 Challenge on both public and private subsets while showing no evidence of overfitting. Our source code is publicly available: \href{https://github.com/v-iashin/CORSMAL}{\UrlFont{github.com/v-iashin/CORSMAL}}

\end{abstract}

\section{Introduction}
In the past, the usage of robots has been confined to heavy industry (e.g. automotive) where the manipulation tasks could be performed in constrained environments with limited human presence. Although this technology was very successful in this setting, it is inadequate for some of the growing areas, such as the light manufacturing industries (e.\,g., food, consumer electronics) and domestic robotics (e.\,g., home assistants and companions).

In order to create robots capable of living and operating in the same space with humans, robots have to be able to interact and co-operate with them.
Manipulating objects of daily living scenarios is an easy cognitive task for humans, who are capable to generalise their grasping and manipulation skills to a wide spectra of materials and shapes. This creates the necessity of developing robots that can cope with such object diversity while performing handover and manipulation tasks. This becomes even more challenging when dealing with previously unseen objects as their dimensions, mass and material are unknown. 
While it is relatively simple for humans to estimate the physical properties (such as mass, stiffness) of the object using vision and other sensing modalities (e.g. tactile, force feedback), it still is an open problem in robotics.

In recent decades, the capabilities of computer vision algorithms have increased significantly due to increased availability in labeled vision data and more performing vision algorithms, based on learnt visual features (such as \cite{he2016deep}).
Computer Vision researchers have put efforts into developing solutions that support robotic manipulation. However, they rely mostly on pre-selected sets of object models (e.g. \cite{peng2019pvnet}, \cite{wang2019densefusion}, \cite{xiang2017posecnn}). For real robotic scenarios, it is important to maintain a certain level of flexibility, reducing the prior knowledge required to deal with the wide spectra of situations required for the collaboration between a human and a robot.
Furthermore, while vision is very effective in localising and estimating the physical properties of solid objects (\cite{phillips2016seeing}, \cite{Wang_2019_CVPR}, \cite{wang2019fast} \cite{hampali2020honnotate}, \cite{sajjan2020clear}, \cite{kokic2019learning}), it is much harder to estimate the filling properties of container objects with different possible types of filling, unless the container is transparent \cite{mottaghi2017see}.
Previous works, such as \cite{liang2019making}, \cite{liang2020robust} and \cite{griffith2012object}, have proved that sound and haptics can be used to estimate the quantity and quality of the filling in a container.
Thus, combining different types of modalities can improve to estimate relevant object features that can support safe and accurate handovers.

In this work, we introduce an approach to estimate the filling mass of a container object using audio and multi-view RGB-D sensing modalities. In particular, we split the problem of estimating the filling mass of an object (container) into 3 tasks. We rely on the CORSMAL Container Manipulation Dataset \cite{CORSMAL} which is designed for this problem. 
\begin{enumerate}[label=$T_{\arabic*}$]
    \item \label{task:filling_type}\textbf{Filling type}. 
    In this task, we predict the type of content present in the container object, if it is not empty. Thus, the possible types of content are: empty, pasta, rice, or water (for drinking glasses and cups). The classification of the filling level is performed combining 2 audio-based classifiers: the first one uses ``classical'' audio features (MFCCs, chromagram, energy, spread, etc.) \cite{giannakopoulos2015pyaudioanalysis} in a random forest classifier, while the second one uses VGGish features \cite{Hershey2017} in a GRU \cite{chung2014empirical} model.  
    \item \label{task:filling_level}\textbf{Filling level}. 
    In this task, we estimate the percentage of the container that is filled with the filling items. The percentage values are discretised into 3 classes: 0, 50, and 90 percent. The classification of this task combines three models: an RGB-based and two audio-based classifiers. The audio-based models are similar to the \ref{task:filling_type} while the RGB-based classifier uses the R(2+1)d features and a GRU model.
    \item \label{task:container_capacity}\textbf{Container capacity}. In this task, we estimate the volume of the container object. Different from the previous tasks, the target value is a real number. 
    For this task, RGB-D + IR data is used to localise the object and estimate its dimensions \cite{2019Xomperomultiview}. The volume is then computed using a cylindrical approximation.
\end{enumerate}
Finally, once estimated the filling type ($T_1$), filling level ($T_2$), and container capacity ($T_3$), we can calculate the mass of the object easily, assuming a pre-estimated average mass density for each filling type. 
The solution that we present resulted the top-ranking one for overall performance among the participants of the CORSMAL Challenge 2020 \cite{2020matilla}. 

The main contributions of this article are the following:
\begin{enumerate}
    \item We present an approach to estimate the container's filling mass based on several different modalities (audio, RGB, IR, and depth) which is formulated as the combination of three sub-tasks: estimation of the filling level of an object, the filling type, and its capacity.
    \item An experiment is performed on a public dataset provided in the context of the CORSMAL Challenge 2020 \cite{2020matilla}. Our model achieves state-of-the-art performance on both public and private test subsets proving generalization capabilities on previously unseen objects.
\end{enumerate}

Please note that this work is conducted in the context of a competition and some decisions are taken under time limitation.

The paper is organized as follows. Section~2 describes our approach to the estimation of the container filling mass and each of the sub-tasks individually as well as the implementation details. The experimentation is presented in Section~3, which includes dataset description, metrics, and results. Section~4 concludes.
\begin{figure}
    \centering
    \includegraphics[width=\textwidth]{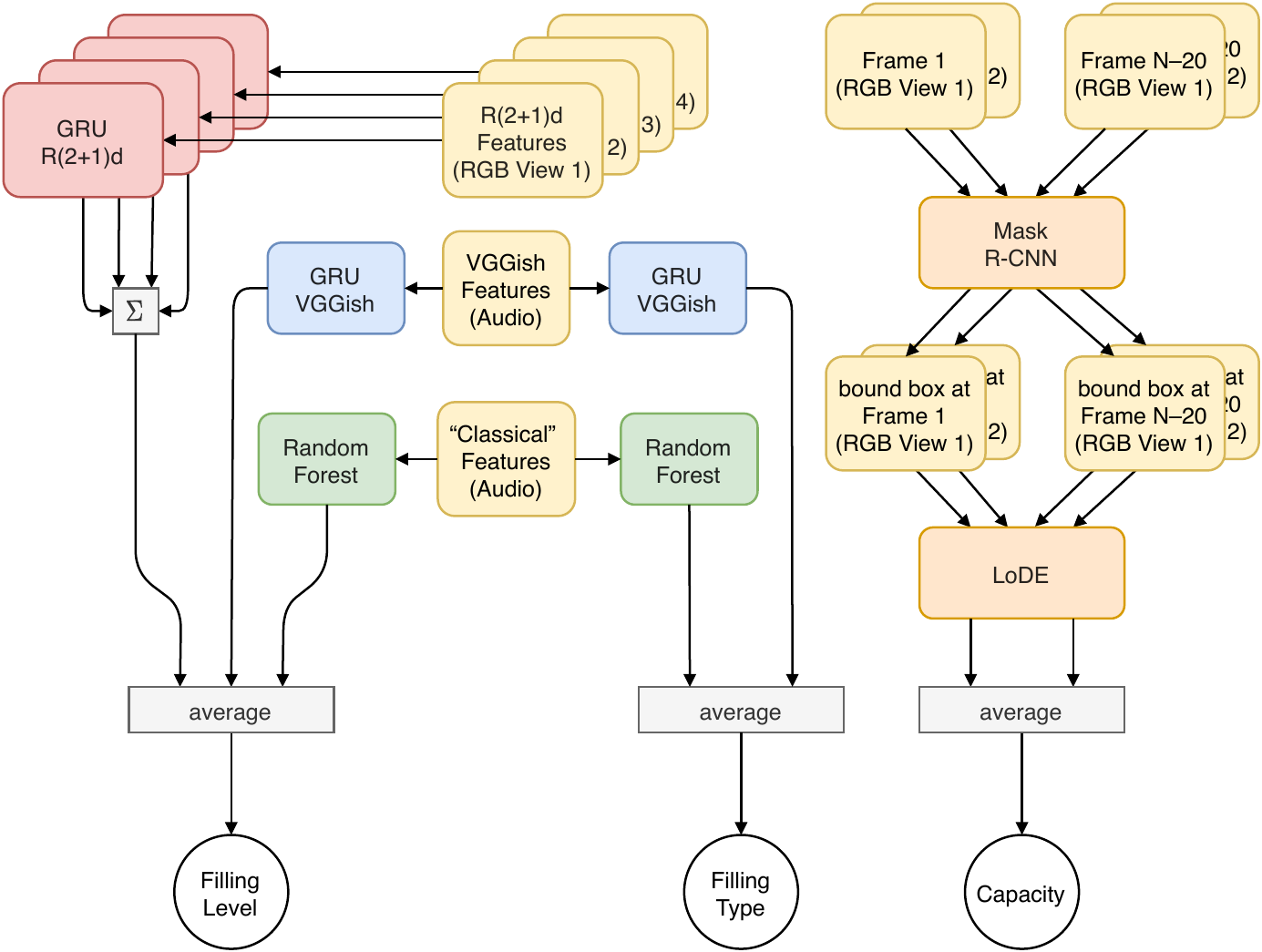}
    \caption{The design of an individual model for each sub-task: container filling level (left), filling type (middle) classification, and container capacity estimation (right). For the \textbf{filling level} estimation we rely on both modalities: RGB (R(2+1)d features) and audio (``classical'' and VGGish features). In particular, we encode streams from each camera placed at different view points with the individual GRU models. Next, we sum up the logits and apply softmax at the output. Next, we process VGGish featured extracted from the audio stream and process it with another GRU model which outputs the class probabilities. Then, the ``classical'' features are classified with the random forest algorithm. Finally, the probabilities from all three models are averaged to produce the final prediction for the filling level. The procedure for \textbf{filling type} classification resembles the one for the filling level except for the absence of RGB stream. The \textbf{capacity estimation} pipeline starts with the extraction of two frames from two camera views (C1 and C2): the 1$^{\text{st}}$ and the 20$^{\text{th}}$ frames to the end. The frames are passed through the Mask R-CNN to detect a container. The detected bounding boxes are used to crop both frames. The crops are sent to the LoDE model which outputs the estimation of the container capacity. If an object was not detected on either of the frames, we use the training prior.}
    \label{fig:main}
\end{figure}

\section{Our Approach}
The task of container mass estimation requires solving three individual sub-tasks: container's filling type and filling level as well as container's capacity estimation. The first two are classification tasks and the latter one is 3D localization. We solve each of the three sub-tasks individually and using the obtained predictions to calculate the mass of a container. We summarize the approach in Figure~\ref{fig:main}.
    
\subsection{Filling Level and Filling Type Estimation}
The first two sub-tasks require to estimate the filling level and filling type of the container. We formulate both sub-tasks as classification problems. We combine the details for both sub-tasks in one section as we are using a similar approach which differs only in the number of output classes. The approach to these two sub-tasks could be roughly divided into three parts: feature extraction, a classification model, and how the predictions are combined if several classifiers were used.
  
\subsubsection{Feature Extraction} 
Being some of the containers not transparent, it would be difficult to rely on vision to recognise the filling type. 
Thus, it can be assumed that a model designed for \textit{filling type} classification should mostly benefit from audio input as a human, if blinded, could distinguish what is being poured by relying on the sound (e.\,g. rice vs. water).
Similarly, for the \textit{filling level}, the audio might provide information if the container was empty or almost full. However, it might also be useful to let a model utilize some information from the vision as some of the containers are transparent and the current \textit{filling level} could be guessed from the visual information. However, for the opaque containers, vision is not informative. Therefore, we rely on audio in both and additionally employ visual modality for the \textit{filling level} task.

To extract audio features for the filling type classification, we rely on both ``classical'' audio features and features from a pre-trained deep learning model (VGGish \cite{Hershey2017}).The ``classical'' features are extracted in two phases. In the first phase, we extract short-term features for each $50$ ms window of the audio. These features include the MFCCs, energy, spectral characteristics, zero crossing rate, chroma vector and deviation \cite{giannakopoulos2015pyaudioanalysis}. Then, these short-term features are used to derive long-term features that are fed to the classifier. Long-term features summarize a larger portion of the audio. These contain the means and standard deviations of multiple short-term features. There are 136-dimensions of features in the final model.

The  features are extracted from every $0.96$ seconds of the original one-channel audio and represented as a 128-d vector obtained from the output of the pre-classification layer. The  features form a tensor of size $ \mathbf{R}^{T_{\text{vgg}} \times 128} $ for each event where $ T_{\text{vgg}} = \text{floor}(T_{\text{sec}} / 0.96) $, $T_{\text{sec}}$ is the duration of the audio in seconds.

For the filling level estimation sub-task we also utilize visual information, which was encoded with a pre-trained model for action recognition (R(2+1)d RGB-only \cite{tran2018closer}). The R(2+1)d features are a 512-d vector extracted from the pre-classification layer which spans about 0.5 second of the original video (16 RGB frames @ 30 fps). Therefore, for each RGB stream obtained from a camera, we extract $ \mathbf{R}^{T_{\text{r21d}} \times 512} $ tensor where $ T_{\text{r21d}} = \text{floor}(T_f / 16) $ with $T_f$ being the total number of frames in a video. 
These specific feature extractors (VGGish and R(2+1)d) as they show promising results not only on the tasks they are trained to solve (sound and action classification) but also when applied for other tasks (\cite{MDVC_2020}, \cite{BMT_2020}, \cite{Liu19a}). Adding these extra features are also motivated with the fact that it is useful to have more diverse individual ``opinions'' (predictions) for a stronger combined prediction \cite{breiman1996bagging}, \cite{king2000better}.

\subsubsection{Classification Models} 
We applied the ``classical'' audio features to both SVM and random forest classifiers \cite{Statistics01randomforests}. We selected the random forest as it performed better on the validation set. The random forest fits a set of decision tree classifiers on random subsets of observations from the dataset as well as the random subset of features. 

Considering that the VGGish and R(2+1)d features are sequence and the expected output of our model is the class label for filling type and filling level this problem can be treated as ``many-to-one'' type. For these types of problems, an RNN is usually considered as a natural candidate. In this work, we tried both LSTM \cite{hochreiter1997long} and GRU \cite{chung2014empirical} but the latter one was preferred due to both efficiency and accuracy.


The last hidden state at the top layer of GRU is used for classification. Specifically, we pass it through one fully-connected layer which maps the hidden state space into the space of labels (for instance, 3 logits for the filling level and 4 for the filling type tasks). Since we may have several RGB sequences (an event is recorded from several cameras) we may aggregate predictions by summing the individual logits obtained from each RGB sequence, e.\,g. R(2+1)d features. Note that we do not take into consideration the fact that if the container is a food box, the filling type could not be liquid.
    
\subsubsection{Post-Processing}
We aggregate the predictions from the models: random forest on ``classical`` audio features, GRU on VGGish features, and, for the filling level, also the GRU output aggregated from multiple cameras on R(2+1)d features. In our work, we use the simple average of probabilities (``opinions'') from each model to form the final prediction for the sub-tasks.

\subsection{Container capacity Estimation}
This task requires the estimation of the capacity of the container. Since the shape and size of a container can vary significantly and the model could be tested in different scenarios, it is a challenging task to tackle.
Our approach builds on the Localisation and Object Dimensions Estimator (LoDE) algorithm \cite{2019Xomperomultiview}. LoDE is a method based on Mask R-CNN \cite{he2017mask} to simultaneously localise objects used as a container and estimate their dimensions. It makes use of RGB frames extracted from the two calibrated RGB cameras positioned on the left and right corners of the room.
The depth and infrared data are also used to improve robustness.

\begin{figure}[t]
    \centering
    \includegraphics[width=\textwidth]{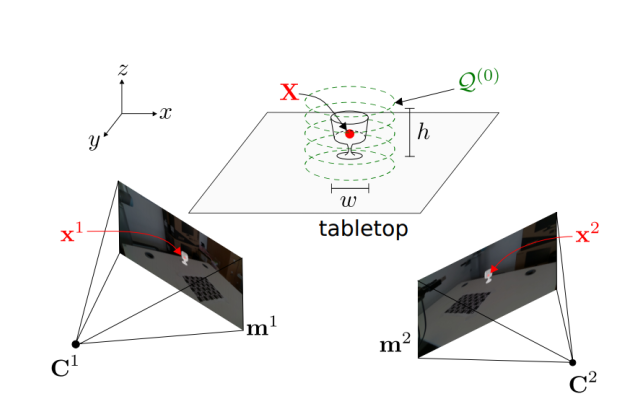}
    \caption{Two frames $m^{1}$ and $m^{2}$ extracted from cameras $C^{1}$ and $C^{2}$ from two different viewpoints (left and right angulation). The 2D centroids, $x^{1}$ and $x^{2}$, are used to triangulate the 3D centroid $X$ of the object. Image published in \cite{2019Xomperomultiview}.}
    \label{fig:LoDEAlgorithm}
\end{figure}

A mask of the object is created using a Mask R-CNN ResNet-50-Feature Pyramid Network (FPN) \cite{huang2017speed} pre-trained on Common Objects in Context (COCO) dataset~\cite{lin2014microsoft}.
The two 2D centroids ($x^{1}$ and $x^{2}$) are estimated in the segmented images from the left and right camera. 
To estimate the 3D centroid ($X$) of an object, a triangulation with the two 2D centroids is performed.
The shape of the object is then predicted by initialising a cylindrical model around the estimated 3D object centroid which iteratively fits the object shape.
The complete algorithm is given in \cite{2019Xomperomultiview}.
Figure \ref{fig:LoDEAlgorithm} \cite{2019Xomperomultiview} shows the above-mentioned method. If the algorithm fails to detect the container, we employ the prior from the training dataset as our prediction.

After the algorithm predicts the width and height of the object, we determine the capacity with the following equation which we designed for the task:
\begin{equation}
    C = \overline{r}^{2} \cdot h \cdot \pi
\end{equation}
where $C$ is the estimated capacity, $\overline{r}$ the average of the radius of the calculated cylindrical shape and $h$ the estimated height of the object. 

\subsection{Implementation Details}
We rely on \texttt{pyAudioAnalysis} library\footnote{\href{https://github.com/tyiannak/pyAudioAnalysis}{\UrlFont{github.com/tyiannak/pyAudioAnalysis}}} to extract the ``classical'' features from the audio and tune the hyper-parameters of the random forest model \cite{giannakopoulos2015pyaudioanalysis}.
VGGish and R(2+1)d features are extracted using \texttt{video\_features}\footnote{\href{https://github.com/v-iashin/video_features}{\UrlFont{github.com/v-iashin/video\_features}}} script. To estimate the capacity of the containers, a modified version of the official LoDE codebase\footnote{\href{https://github.com/CORSMAL/LoDE}{\UrlFont{github.com/CORSMAL/LoDE}}} was used.

We tune only the \emph{number of trees} parameter of the random forest classifier. The other parameters are kept as the default values. We retrain the model with different values of this parameter and choose the value that achieves the highest accuracy in the validation set.
To train the GRU models we employ batching of size 64 while padding the shorter sequences in a batch to the length of the longest sequence. The sizes of the hidden states are 512. The GRU has 5 layers for VGGish features and 3 layers for R(2+1)d features. We train the GRU models for up to 30 epochs with Adam optimizer \cite{kingma2014adam} with learning rate $3\cdot 10^{-4}$ and cross-entropy loss. These hyper-parameters are selected on the 3-fold cross-validation which is described next.

Since there is uncertainty about the objects which will occur during the inference, a model requires to be capable to generalize its predictions to unexpected forms and sound feedback of containers. To prevent our model from overfitting, we employ 3-fold cross-validation. 

The splits are constructed following the distribution of the hold-out test set. Specifically, we split ``train'' objects according to their type 2:1 such that the validation dataset will have one object and the other two are used for training. Therefore, we had 3 train:validation splits with 6:3 objects (if 9 objects are in the training set) in each such that each object appears in a validation set once.
    
Running the training and evaluation scripts are executed on one 1080Ti and takes about 8 hours with R(2+1)d feature extraction allocates more than half of the execution time. This time can be significantly reduced by increasing the number of GPUs. We provide the \texttt{Docker}\footnote{\href{https://hub.docker.com/r/iashin/corsmal}{\UrlFont{hub.docker.com/r/iashin/corsmal}}} image to reproduce the results and the source code\footnote{\href{https://github.com/v-iashin/CORSMAL}{\UrlFont{github.com/v-iashin/CORSMAL}}}.
\begin{table}[t]
\centering
\setlength{\tabcolsep}{5pt}
\begin{tabular}{lrrrrrr}
\toprule
  & \multicolumn{3}{c}{\textbf{Validation Set}} & ~ & \multicolumn{2}{c}{\textbf{Test Set}} \\ \cmidrule{2-7} 
\textbf{Sub-task} & \textbf{``class.'' feats.} & \textbf{VGGish} & \textbf{R(2+1)d} & ~ & \textbf{Public} & \textbf{Private} \\ \midrule
Filling Level & 69.9 & 75.5 & 74.7 & ~ & 78.14 & 81.16 \\
Filling Type & 93.3 & 91.3 & -- & ~ & 93.83 & 94.70 \\\bottomrule
\end{tabular}
\vspace{1em}
\caption{Performance of individual classification models for the container's filling level and filling type sub-tasks. The results are shown using the average F1-measure obtained by averaging the scores obtained with 3-fold cross-validation and the values obtained from the CORSMAL Challenge 2020 leader board. \label{tab:val}}
\end{table}

\section{Experiments}

\subsection{Dataset} We train our model on the official training set of the challenging audio-visual-inertial CORSMAL dataset \cite{CORSMAL}. The CORSMAL dataset is a collection of human interaction with different containers recorded with multi-sensor devices: 8-element circular microphone and 4 cameras from different viewpoints recording RGB, IR, depth, IMU streams.

The dataset consists of three parts: public train, public test, and private test. The public train holds 9 containers: 3 plastic drinking cups, 3 drinking glasses, and 3 cardboard food boxes (e.\,g. cereal); 1 cup, 1 glass, and 1 food box is reserved for the public test set. The private test is hidden and known to have 3 more containers. The difference between the public and private test sets is in the absence of information about the input. 

A container can be filled with rice, pasta, or water (only glasses and cups) and at 3 different levels: 0, 50, and 90\,\% with respect to the capacity of the container. All combinations of containers are executed by 12 different subjects for 2 different backgrounds and 2 different illumination conditions. In total, there are 1140 different combinations.
    
\subsection{Metrics} To track the performance of our classification algorithms, we employ the F1-score which favors a balance between precision and recall. The metric is weighted to account for the class imbalance. While the modified absolute error was used in evaluating container capacity. For a more detailed explanation of how the metrics are calculated on the public and private subsets of CORSMAL, we refer a reader to the technical document provided by CORSMAL Challenge 2020.
    
\subsection{Results}
\begin{table}[t]
\centering
\setlength{\tabcolsep}{5pt}
\begin{tabular}{l r r}
\toprule
                             & \multicolumn{2}{c}{\textbf{Test set}} \\ \cmidrule{2-3} 
\textbf{Sub-task}                & \textbf{Public}   & \textbf{Private}  \\ \midrule
Filling Level                & 78.14             & 81.16             \\
Filling Type                 & 93.83             & 94.70             \\
Container Capacity           & 60.56             & 60.58             \\ \midrule
Overall Performance          & 64.98             & 65.15               \\\bottomrule
\end{tabular}
\vspace{1em}
\caption{The results from the CORSMAL Challenge 2020 leader board by November 23rd 2020. The performance is shown using the modified absolute error between the estimated filling mass and the ground truth. \label{tab:test}}
\end{table}

\subsubsection{Container Filling Level Classification} 
We compare the performance of the container's filling level between each classification model individually. Table~\ref{tab:val} (first row) shows the results of the classification models averaged among three validation folds on the F1 metric. According to the results, the performance of the VGGish-based GRU model (audio) performs on par with the GRU model which is based on R(2+1)d (visual) features (75.5 and 74.7) while the random forest-based on ``classical'' features performs the worst (69.9), yet adding it to the final combination improves the overall performance. 
    
The results of the combination of three classification models show the importance of aggregation of the individual predictions as the results on both public and private sets are higher than of the individual models. We also highlight the fact that according to the difference between private and public scores our model has not shown any sign of over-fitting but rather a strong generalization capability.
    
\subsubsection{Container Filling Type Classification}
Similar to the filling level sub-task, the classification results of filling type models are compared individually (see Table~\ref{tab:val}, the second row). The results show the importance of audio modality and ``classical'' features, in particular. The random forest classifier achieves 93.3 F1 score which slightly higher than VGGish-based GRU with 91.3 F1 score.\footnote{We also trained the GRU model on R(2+1)d features, yet it reached only F1=67.3.}
    
For this sub-task, we also observe the improvement from combining the predictions. Yet, fewer gains are obtained in performance possibly because we rely on only one modality (audio). Nevertheless, the model also preserves the generalization capabilities and absence of over-fitting.

\subsubsection{Container Capacity Estimation}
For the prediction of the container capacity, a modified version of the LoDE method is implemented.
The original LoDE algorithm was developed for calculating the capacity of transparent objects and was tested on a supervised environment with no occluded objects.
For the CORSMAL challenge, the objects were manipulated by humans with multiple occlusions and additional items in the camera view.

The three target objects for the public set of the challenge are a beer cup, a cocktail glass, and a pasta box.
Since the original Mask R-CNNs were pre-trained on the COCO dataset, there was no label to recognise a pasta box or other similar boxes. To alleviate this, an additional label was introduced.
The label is equal to the book (present in the COCO dataset), considering the similarity of its shape to a box.
For the future, transfer learning will be considered to better adjust the final weight of the fully connected layer of the Mask R-CNN and better segment the objects.

In the dataset, some videos start with an empty room (no object or person in it).
To avoid having a frame with no object to detect, two frames were extracted from the videos: the first and the $20^{th}$ to last frame.
An average of the capacities detected in the two frames was then used for final validation.

\begin{figure}
    \centering
    \includegraphics[width=\textwidth]{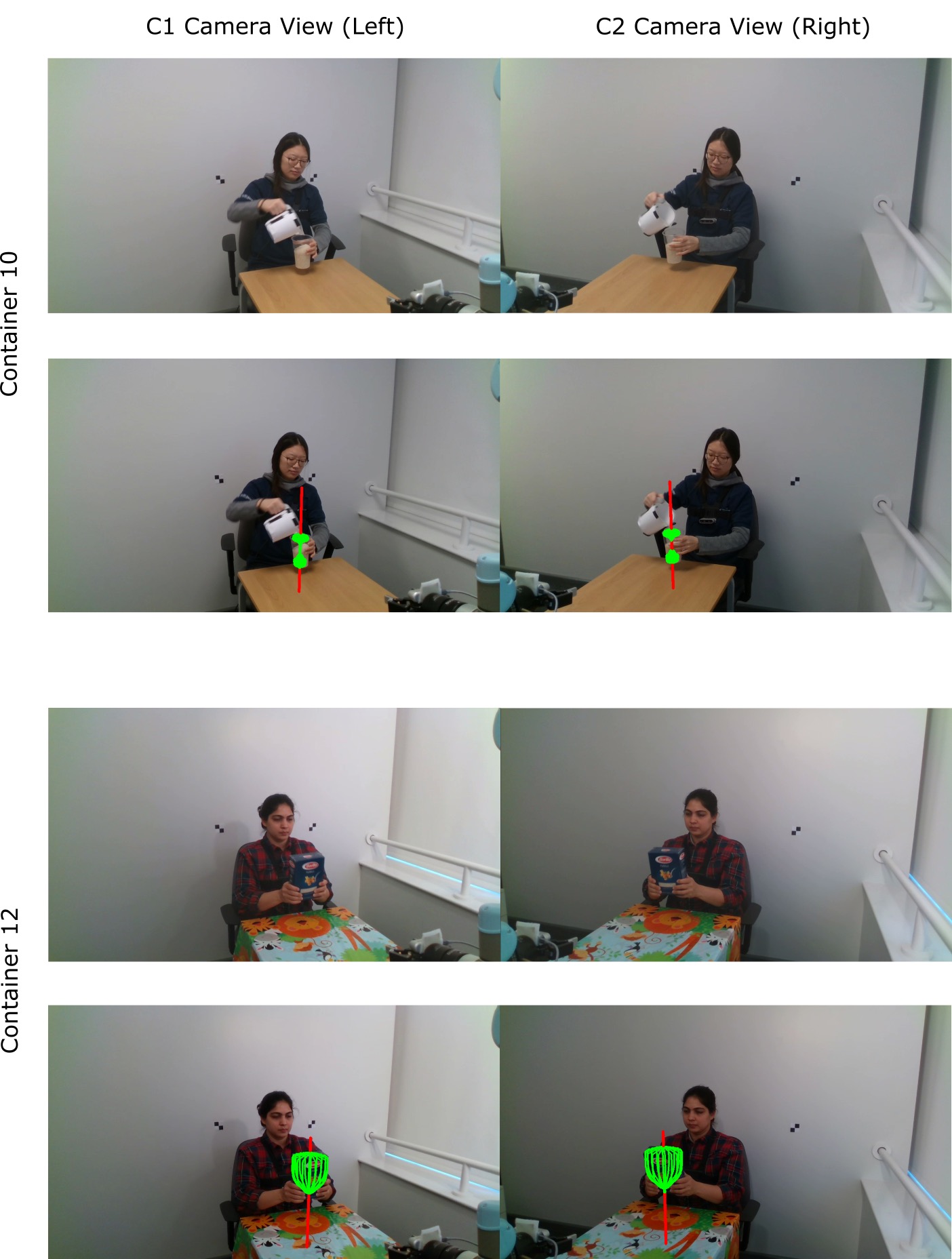}
    \caption{Examples showing the limitations of the capacity estimation of two of the objects (10 and 12) of the CORSMAL Challenge. The object 10 is occluded by the hand and the object 12 is predicted to be the wrong object, because the training set did not contain boxes. }
    \label{fig:task3_results}
\end{figure}

Figure \ref{fig:task3_results} shows two examples of the resulting capacity estimation through LoDE estimation. For the beer cup (object 10), occlusion by the human hand breaks the mask into two. 
Thus, predicting a smaller capacity for the container than the real one.
For the pasta box (object 12), the model correctly identifies the container object, although not trained for it.
The calculated capacity, similarly to the beer cup, is smaller than the ground-truth, because of the occlusion by the manipulator's hands.
Nevertheless, the model achieves an accuracy of 60.56\% when tested on the public dataset and 60.58\% on the private dataset, for an overall of 60.58\%, which is higher than the average (see Table~\ref{tab:test}).

\subsubsection{Container Filling Mass Estimation (the overall task)}
\begin{figure}[t]
    \centering
    \includegraphics[width=\textwidth]{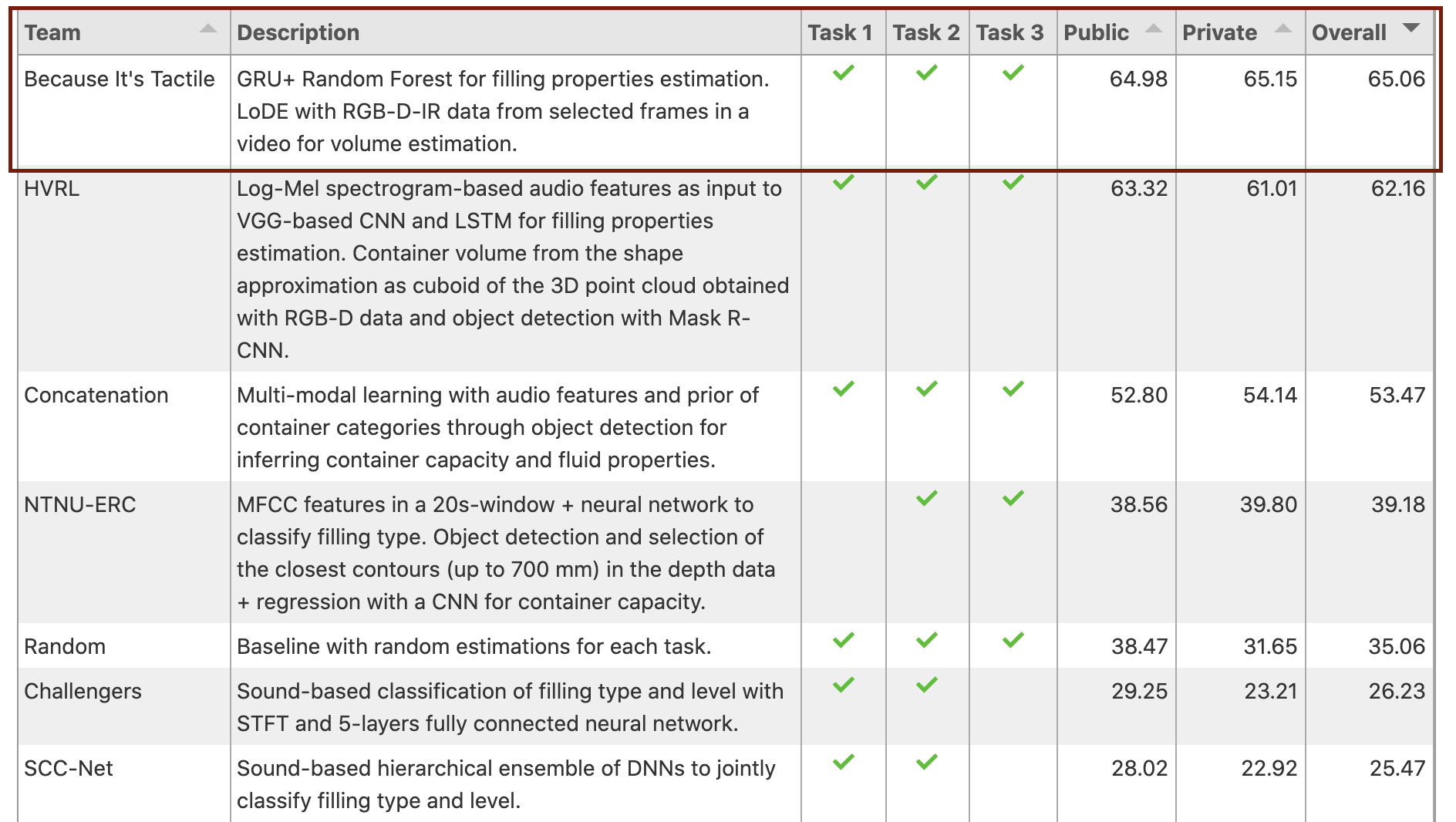}
    \caption{The state of the leader board on 22$^\text{th}$ of November 2020 showing our team (``Because It's Tactile'') placed first in the overall task (filling mass estimation) \cite{corsmalwebsite}.}
    \label{fig:lb}
\end{figure}

Table~\ref{tab:test} (bottom row) shows the results of the container filling mass estimation which takes into consideration the performance of all three sub-tasks together. Since the overall performance metric favours the least ``performing'' sub-task in the final score we believe it could be further improved by exploring the container capacity estimation sub-task. Our approach outperforms all other submissions to CORSMAL Challenge 2020 leaving us on the top of the leader board (see Figure~\ref{fig:lb}).

\section{Conclusion}
Complex real-world sensing tasks as those posed in the CORSMAL 2020 Challenge require analysis of multiple data modalities. The amount of information gained by a modality is dependent on the nature of the task and it may be predicted by human intuition. However, a modality may still improve the combined performance, despite having poor performance alone. We observed this in our experiments. We predicted audio modality to be informative in filling level and type classification. Indeed, it achieves decent performance. However, the addition of visual features improves overall accuracy. Besides different modalities, using multiple models that follow different paradigms on the same modality increased the accuracy of our predictions. Achieving the highest score in this challenge was possible thanks to the variety in both the modalities and the algorithms. 

In future works, other modalities such as IMU and tactile sensing can be explored. Also, the capacity estimation algorithm will be further improved by training the model for the Mask-RCNN which will permit to recognize objects more reliably.
In addition, the LoDE model will be ulterior improved by developing optimal capacity evaluation calculations based on the label of the recognised object.

\bibliographystyle{splncs04}
\bibliography{icpr2020}

\end{document}